\def\year{2021}\relax
\documentclass[letterpaper]{article} 
\usepackage{aaai21}  
\usepackage{times}  
\usepackage{helvet} 
\usepackage{courier}  
\usepackage[hyphens]{url}  
\usepackage{graphicx} 
\usepackage{graphicx}  
\usepackage{natbib}  
\usepackage{caption} 
\usepackage{amsmath}
\usepackage{amssymb}
\usepackage{amsthm}
\usepackage{algorithm}
\usepackage{algpseudocode}

\usepackage{bm}
\usepackage{booktabs}

\usepackage{capt-of}
\usepackage{comment}
\usepackage{courier}
\usepackage{color}
\usepackage{cite}
\usepackage[figurename=Fig]{caption}
\usepackage{comment}
\usepackage[justification=raggedright]{caption}

\usepackage{enumitem}
\usepackage{etoolbox}

\usepackage{graphicx}
\usepackage{helvet}

\usepackage[utf8]{inputenc}
\usepackage{leftidx}
\usepackage{multicol}
\usepackage{multirow}

\usepackage{soul}
\usepackage{subcaption}
\usepackage{soul}
\usepackage[para,online,flushleft]{threeparttable}
\usepackage{times}
\usepackage{url}
\usepackage{xcolor}

\title{Multidimensional Uncertainty-Aware Evidential Neural Networks}

\author{Yibo Hu\textsuperscript{\rm 1}, Yuzhe Ou\textsuperscript{\rm 1}, Xujiang Zhao\textsuperscript{\rm 1},  Jin-Hee Cho\textsuperscript{\rm 2}, Feng Chen\textsuperscript{\rm 1}\\
}

\affiliations{
    \textsuperscript{\rm 1} Department of Computer Science, The University of Texas at Dallas, Richardson, TX, USA\\ 
    \textsuperscript{\rm 2} Department of Computer Science, Virginia Tech, Falls Church, VA, USA\\ 
    \{yibo.hu, yuzhe.ou, xujiang.zhao\}@utdallas.edu, jicho@vt.edu, feng.chen@utdallas.edu
}

\begin{document}

\maketitle

\begin{abstract}

Traditional deep neural networks (NNs) have significantly contributed to the state-of-the-art performance in the task of classification under various application domains.  However, NNs have not considered inherent uncertainty in data associated with the class probabilities where misclassification under uncertainty may easily introduce high risk in decision making in real-world contexts (e.g., misclassification of objects in roads leads to serious accidents).  Unlike Bayesian NN that indirectly infer uncertainty through weight uncertainties, evidential NNs (ENNs) have been recently proposed to explicitly model the uncertainty of class probabilities and use them for classification tasks.  An ENN offers the formulation of the predictions of NNs as subjective opinions and learns the function by collecting an amount of evidence that can form the subjective opinions by a deterministic NN from data.  However, the ENN is trained as a black box without explicitly considering inherent uncertainty in data with their different root causes, such as vacuity (i.e., uncertainty due to a lack of evidence) or dissonance (i.e., uncertainty due to conflicting evidence).  By considering the multidimensional uncertainty, we proposed a novel uncertainty-aware evidential NN called {\em WGAN-ENN (WENN)} for solving an out-of-distribution (OOD) detection problem.  We took a hybrid approach that combines Wasserstein Generative Adversarial Network (WGAN) with ENNs to jointly train a model with prior knowledge of a certain class, which has high vacuity for OOD samples.  Via extensive empirical experiments based on both synthetic and real-world datasets, we demonstrated that the estimation of uncertainty by WENN can significantly help distinguish OOD samples from boundary samples. WENN outperformed in OOD detection when compared with other competitive counterparts.

\end{abstract}

\section{Introduction} \label{sec:intro}
Deep Learning (DL) models have recently gained tremendous attention in the data science community.
Despite their superior performance in various decision making tasks, inherent uncertainty derived from data based on different root causes has not been sufficiently explored.  Predictive uncertainty estimation using Bayesian neural networks (BNNs) has been explored for classification prediction or regression in computer vision applications~\cite{kendall2017uncertainties}. They considered well-known uncertainty types, such as aleatoric uncertainty (AU) and epistemic uncertainty (EU), where AU only considers data uncertainty caused by statistical randomness (e.g., observation noises) while EU refers to model uncertainty introduced by limited knowledge or ignorance in collected data.  On the other hand, in the belief/evidence theory, Subjective Logic (SL)~\cite{josang2018uncertainty} considered vacuity, which is caused by a lack of evidence, as the key dimension of uncertainty.
In addition to vacuity, they also defined other types of uncertainty, such as dissonance (e.g., uncertainty due to conflicting evidence) or vagueness (e.g., uncertainty due to multiple beliefs on a same observation).

Although conventional deep NNs (DNNs) have been commonly used to solve classification tasks, uncertainty associated with classification classes has been significantly less considered in NNs even if the risk introduced by misclassification may bring disastrous consequence in real-world situations, such as car crash due to the misclassification of objects in roads.  Recently, techniques using evidential neural networks (ENNs)~\cite{sensoy2018evidential} have been proposed to explicitly model the uncertainty of class probabilities.  An ENN uses the predictions of an NN as subjective opinions and learns a function that collects an amount of evidence to form the opinions by a deterministic NN from data.  However, the ENN is trained as a black box without explicitly considering different types of uncertainty in the data (e.g., vacuity or dissonance), which often results in overconfidence when tested with out-of-distribution (OOD) samples. We measure the extent of confidence in a given classification decision based on the high class probability of a given class (i.e., a belief in SL).  Overconfidence refers to a high class probability in an incorrect class prediction.  To mitigate the overconfidence issue, regularization methods have proposed to hand-pick auxiliary OOD samples to train the model~\cite{malinin2018predictive, zhao2019quantifying}.  However, the regularization methods with prior knowledge require a large amount of OOD samples to ensure the good generalization of a model behaviour to the whole data space.  

\begin{figure}[!htbp]
\centering
\includegraphics[width=0.47\textwidth]{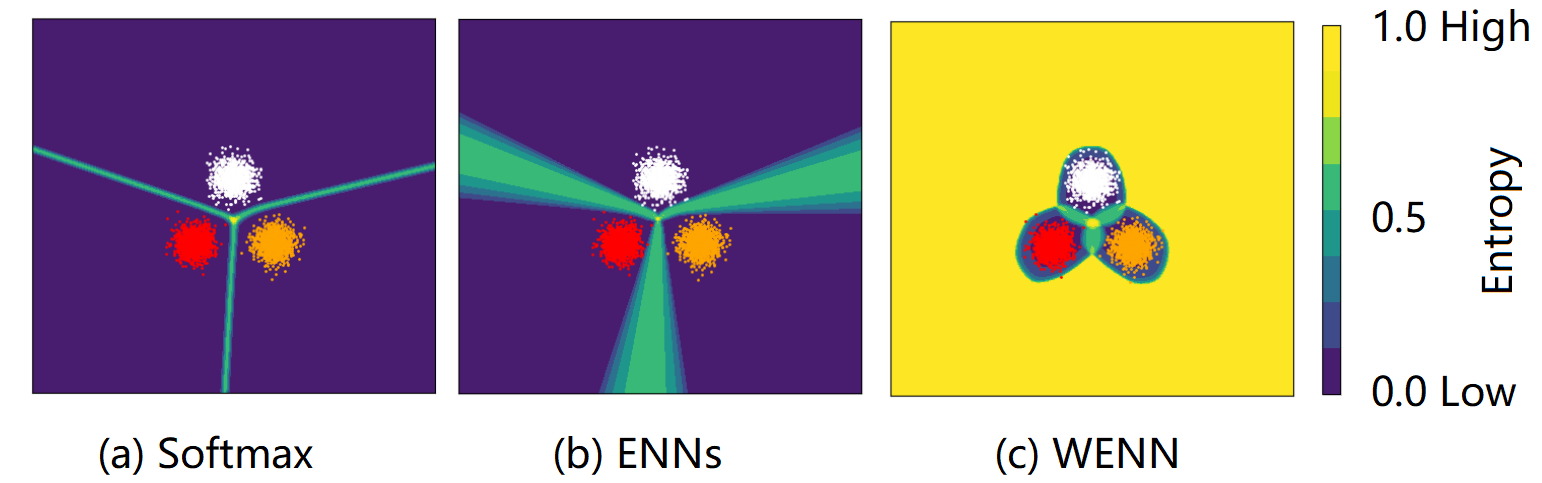}
\caption{Uncertainty (entropy) estimation based on synthetic data: (a) using standard NNs with softmax function; (b) using ENNs; 
and (c) using WENN. Only (c) shows high uncertainty in both boundary and OOD regions.
} \label{fig1: example1}
\end{figure}

In this work, we propose a model called {\em WGAN-ENN (WENN)} that combines ENNs with Wasserstein GAN (WGAN)~\cite{arjovsky2017wasserstein} to jointly train a model with prior knowledge of a certain class (e.g., high vacuity OOD or high dissonance in-class boundary samples) to reinforce and achieve high prediction confidence only for in-distribution (ID) regions, high vacuity only for OOD samples, and high dissonance only for ID boundary samples.  

To briefly demonstrate the performance of uncertainty estimation by conventional NNs, ENNs, and our proposed WENN model, we explain it with a simple three-class classification problem in Fig~\ref{fig1: example1}. 
We measure entropy~\cite{shannon1948mathematical} estimated on the predictive class probabilities by three different approaches: 
(i) Fig~\ref{fig1: example1}~(a) shows the prediction for boundary or OOD samples by traditional NNs (i.e., BNN to indirectly infer uncertainty through weight uncertainties) using the softmax and demonstrates overconfidence; (ii)  Fig~\ref{fig1: example1}~(b) shows the overconfidence in the prediction of OOD samples by the ENN; 
and (iii) Fig~\ref{fig1: example1} (c) shows the high confidence in the prediction of the ID region by WENN.

This work provides the following {\bf key contributions}:
\begin{itemize} 
\item We considered inherent uncertainties derived from different root causes by taking a hybrid approach that leverages both deep learning and belief model (i.e., Subjective Logic or SL). 
Although both fields have studied uncertainty-aware approaches to tackle various kinds of decision making problems, there has been lack of efforts to leverage both of their merits. 
We believe this work sheds light on the direction of incorporating both fields.

\item We considered ENNs to quantify multidimensional uncertainty types in data and learn subjective opinions. In particular, the subjective opinions formulated by the SL can be easily leveraged for the quantification of multidimensional uncertainties where we measured {\em vacuity} and {\em dissonance} based on SL. 

\item Our proposed WENN, combining WGAN and ENNs, can generate a sufficient amount of auxiliary OOD samples for training and use the Wasserstein distance to measure the variety of those samples.  Our proposed alternating algorithm can leverage all the intermediate samples more efficiently than other regularized methods. 

\item We demonstrated that WENN outperforms competitive state-of-the-art counterparts in OOD detection, showing 7\% better performance than the best of the counterparts in the most difficult scenario CIFAR10 vs CIFAR100.

\end{itemize}

\section{Related Work} \label{sec:related-work}

\noindent {\bf Uncertainty Quantification in Bayesian Deep Learning (BDL)}: Machine/deep learning (ML/DL) researchers considered {\it aleatoric} uncertainty (AU) and {\it epistemic} uncertainty (EU) based on Bayesian Neural Networks (BNNs) for computer vision applications.  AU consists of homoscedastic uncertainty (i.e., constant errors for different inputs) and heteroscedastic uncertainty (i.e., different errors for different inputs)~\cite{gal2016uncertainty}. A BDL framework was presented to estimate both AU and EU simultaneously in regression settings (e.g., depth regression) and classification settings (e.g., semantic segmentation)~\cite{kendall2017uncertainties}.   {\em Dropout variational inference}~\cite{gal2016dropout} was proposed as one of key approximate inference techniques in BNNs ~\cite{blundell2015weight,pawlowski2017implicit}. Later {\em distributional uncertainty} is defined based on distributional mismatch between the test and training data distributions~\cite{malinin2018predictive}. 
\\

\noindent {\bf Uncertainty Quantification in Belief/Evidence Theory}: In belief/evidence theory, uncertainty reasoning has been substantially explored in Fuzzy Logic~\cite{de2018intelligent}, Dempster-Shafer Theory (DST)~\cite{sentz2002combination}, or Subjective Logic (SL)~\cite{josang2016subjective}.  Unlike the efforts in ML/DL above, belief/evidence theory focused on reasoning of inherent uncertainty in information resulting from unreliable, incomplete, deceptive, and/or conflicting evidence.  SL considered uncertainty in subjective opinions in terms of {\em vacuity} (i.e., a lack of evidence) and {\em vagueness} (i.e., failure of discriminating a belief state)~\cite{josang2016subjective}. Recently, other dimensions of uncertainty have been studied, such as {\em dissonance} (due to conflicting evidence) and {\em consonance} (due to evidence about composite subsets of state values)~\cite{josang2018uncertainty}. In DNNs, \cite{sensoy2018evidential} proposed ENN models to explicitly modeling uncertainty using SL. However, it only considered predictive entropy to qualify uncertainty.
\\

\noindent {\bf Out-of-Distribution Detection}: 
Recent OOD detection approaches began to use NNs in a supervised fashion that outperformed traditional models, such as kernel density estimation and one-class support vector machine in handling complex datasets~\cite{hendrycks2016baseline}. Many of these models \cite{liang2017enhancing, hendrycks2018deep} integrated auxiliary datasets to adjust the estimated scores derived from prediction probabilities. 
These OOD detection models were specifically designed to detect OOD samples. In addition, well-designed uncertainty estimation models are leveraged for OOD detection.  Recently, uncertainty models~\cite{sensoyuncertainty} have shown their preliminary results on OOD detection by conducting performance comparison of various OOD detection models.

\section{Preliminaries} \label{sec:background}
This section provides the background knowledge to understand this work, including: (1) subjective opinions in SL; (2) uncertainty characteristics of subjective opinion; and (3) ENNs to predict subjective opinions. 

\subsection{Subjective Opinions in SL}
A multinomial opinion in a given proposition $x$ is represented by $\omega_Y = (\bm{b}_Y, u_Y, \bm{a}_Y)$ where a domain is $\mathbb{Y} \equiv \{1, \cdots, K\}$, a random variable $Y$ takes value in $\mathbb{Y}$ and $K = |\mathbb{Y}| \geq 2$. The additivity requirement of $\omega_Y$ is given as $\sum_{y \in \mathbb{Y}} \bm{b}_Y(y) + u_Y = 1$.  Each parameter indicates,
\begin{itemize}
\item $\bm{b}_Y$: {\em belief mass function} over $\mathbb{Y}$;
\item $u_Y$: {\em uncertainty mass} representing {\em vacuity of evidence};
\item $\bm{a}_Y$: {\em base rate distribution} over $\mathbb{Y}$, with $\sum_y \bm{a}_Y(y)=1$.
\end{itemize}

The projected probability distribution of a multinomial opinion is given by:
\begin{equation} \label{eq:multinomial-projected}
\small
\mathbf{p}_Y(y) = \bm{b}_Y(y) + \bm{a}_Y(y) u_Y,\;\;\; \forall y \in \mathbb{Y}.
\end{equation}  
Multinomial probability density over a domain of cardinality $K$ is represented by the $K$-dimensional Dirichlet PDF where the special case with $K=2$ is the Beta PDF as a binomial opinion. Denote a domain of $K$ mutually disjoint elements in $\mathbb{Y}$ and $\alpha_Y$ the strength vector over $y \in \mathbb{Y}$ and ${\bf p}_Y$ the probability distribution over $\mathbb{Y}$. Dirichlet PDF with ${\bf p}_Y$ as $K$-dimensional variables is defined by:
\begin{eqnarray} \label{eq:multinomial-dir}
\small
\mathrm{Dir}(\bm{p}_Y; {\bm \alpha}_Y) = \frac{1}{B({\bm \alpha}_Y)} \prod_{y \in \mathbb{Y}} \bm{p}_Y (y) ^{({\bm \alpha}_Y(y)-1)},
\end{eqnarray} 
where $\frac{1}{B({\bm \alpha}_Y)} = {\Gamma \Big(\sum_{y \in \mathbb{Y}} {\bm \alpha}_Y (y)\Big)} / {\prod_{y \in \mathbb{Y}} ({\bm \alpha}_Y (y))}$, ${\bm \alpha}_Y(y) \geq 0$, and ${\bf p}_Y (y) \neq 0$ if ${\bm \alpha}_Y (y) < 1$.

We term \textit{evidence} as a measure of the amount of supporting observations collected from data in favor of a sample to be classified into a certain class. Let ${\bf r}_Y(y) \ge 0 $ be the evidence derived for the singleton $y\in \mathbb{Y}$.  The total strength ${\bm \alpha}_Y(y)$ for the  belief of each singleton $y \in \mathbb{Y}$ is given by: 
\begin{eqnarray} \label{eq:multinomial-alpha}
\small
{\bm \alpha}_Y(y) = \bm{r}_Y(y) + \bm{a}_Y(y) W, 
\end{eqnarray}
where $W$ is a non-informative weight representing the amount of uncertain evidence and $\bm{a}_Y(y)$ is the base rate distribution.  Given the Dirichlet PDF, the expected probability distribution over $\mathbb{Y}$ is:
\begin{equation} \label{eq:multinomial-expected}
\small
\mathbb{E}_Y(y) = \frac{{\bm \alpha}_Y (y)}{\sum_{y_i \in \mathbb{Y}} {\bm \alpha}_Y (y_i)} = \frac{\bm{r}_Y(y) + \bm{a}_Y(y) W}{W + \sum_{y_i \in \mathbb{Y}} \bm{r}_Y(y_i)},   \forall_y \in \mathbb{Y}.
\end{equation}
The observed evidence in the Dirichlet PDF can be mapped to the multinomial opinions by:
\begin{equation} \label{eq:multinomial-belief}
\small
\bm{b}_Y(y) = \frac{\bm{r}(y)}{S}, \;
u_Y = \frac{W}{S},  
\end{equation}
where $S = \sum_{y_i \in \mathbb{Y}} {\bm \alpha}_Y(y_i)$. 
We set the base rate $\bm{a}_Y(y) = \frac{1}{K}$ and the non-informative prior weight $W = K$, and hence $\bm{a}_Y(y)\cdot  W = 1$ for each $y \in \mathbb{Y}$, as these are default values considered in subjective logic.

\subsection{Uncertainty Characteristics of Subjective Opinions}

The multidimensional uncertainty dimensions of a subjective opinion based on the formalism of SL are discussed in ~\cite{josang2018uncertainty}. As we deal with a multinomial opinion in this work, we discuss two main types of uncertainty dimensions, which are {\em vacuity} and {\em dissonance}. 
 
\begin{figure}[!htbp]
\centering
\includegraphics[width=0.47\textwidth]{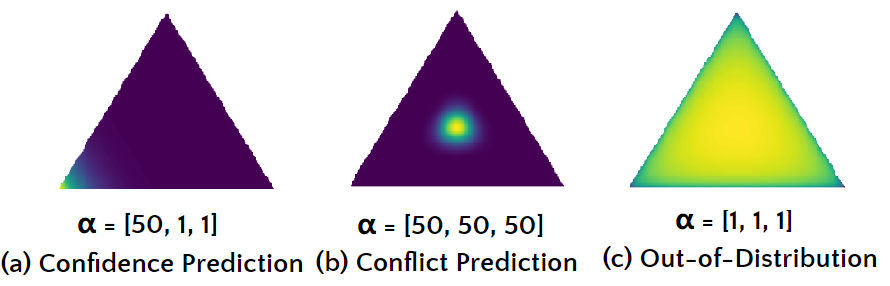}
\caption{Illustration of different vacuity and dissonance of subjective opinions based on their evidence measures. 
} \label{fig:Dirichlet}
\end{figure}

The main cause of {\em vacuity} (a.k.a. ignorance) is a lack of evidence or knowledge, which corresponds to uncertainty mass, $u_Y$, of an opinion in SL as:
\begin{eqnarray}
\small
\text{Vac}({\bm \alpha}_Y) = \frac{W}{S}. \label{eq:vacuity}
\end{eqnarray}
This type of uncertainty refers to uncertainty caused by insufficient information or knowledge to understand or analyze a given opinion.

The {\em dissonance} of an opinion can happen when there is an insufficient amount of evidence that can clearly support a particular belief. 
For example, when a same amount of evidence is supporting multiple extremes of beliefs, high dissonance is observed. Hence, the dissonance is estimated by the difference between singleton belief masses (e.g., class labels), leading to `inconclusiveness' in decision making situations.

Given a multinomial opinion with non-zero belief masses, the measure of dissonance can be obtained by:
\begin{equation}
\small
\label{eq:belief-dissonance-multi}
\text{Diss}({\bm \alpha}_Y) = \sum\limits_{y_{i}\in \mathbb{Y}}\left(\frac{\bm{b}_{Y}(y_{i})\!\!\!\! \sum\limits_{y_j \in \mathbb{Y}\setminus y_i}\!\!\!\!\!\bm{b}_{Y}(y_{j}) \mbox{Bal}(y_{j},y_{i})}{\sum\limits_{y_j \in \mathbb{Y}\setminus y_i}\bm{b}_{Y}(y_{j})}  \right),
\end{equation} 
where the relative mass balance between a pair of belief masses $\bm{b}_{Y}(y_{j})$ and $\bm{b}_{Y}(y_{i})$ is expressed by:
\begin{align}
\small
\label{eq:belief-balance}
\text{Bal}(y_j,y_i)=
  \begin{cases} 
  1-\frac{|\bm{b}_{Y}(y_{j})-\bm{b}_{Y}(y_{i})|}{\bm{b}_{Y}(y_{j})+\bm{b}_{Y}(y_{i})}, 
  & \text{if $\bm{b}_{Y}(y_{j}) \bm{b}_{Y}(y_{i}) \neq 0$}\\
  0, & \text{otherwise.}
  \end{cases}
\end{align}

The above two uncertainty measures (i.e., vacuity and dissonance) can be interpreted using class-level evidence measures of subjective opinions. 
As in Fig.~\ref{fig:Dirichlet}, given three classes (1, 2, and 3), we have three subjective opinions $\{{\bm \alpha}_1, {\bm \alpha}_2, {\bm \alpha}_3\}$, represented by the three-class evidence measures as: ${\bm \alpha}_1 = (50, 1, 1)$ representing low uncertainty (entropy, dissonance and vacuity) which implies high certainty (often represented as high confidence in a decision making context), ${\bm \alpha}_2 = (50, 50, 50)$ indicating high inconclusiveness due to high conflicting evidence which gives high entropy and high dissonance, ${\bm \alpha}_3 = (1, 1, 1)$ showing the case of high vacuity which is commonly observed in OOD samples. 
Based on our observations from Fig~\ref{fig:Dirichlet} (b) and (c), we found that entropy cannot distinguish uncertainty due to vacuity or dissonance, which naturally results in inability to distinguish boundary samples from OOD samples. However, vacuity can effectively detect OOD samples because the cause of uncertainty is from a lack of evidence.

\subsection{Evidential Neural Networks (ENNs)}
ENNs \cite{sensoy2018evidential} are similar to classical NNs except that the softmax layer is replaced by an activation layer (e.g., ReLU) to ascertain
non-negative output, which is taken as the evidence vector for the predicted Dirichlet distribution. 
Given sample $i$, let $f({\bf x}_i | \Theta)$ represents the evidence vector predicted by the network for the classification, where ${\bf x}_i \in \mathbb{R}^L$ is the input feature and $\Theta$ is the network parameters. Then, the corresponding Dirichlet distribution has
parameters ${\bm \alpha}_i = f({\bf x}_{i} | \Theta) + 1$. 
Let ${\bf y}_i$ be the ground-truth label, the Dirichlet density $\text{Dir}({\bf p}_i ; {\bm \alpha})$ is the prior on the Multinomial distribution, $\text{Multi}({\bf y}_i | {\bf p}_i)$. The following sum of squared loss is used to estimate the parameters ${\bm \alpha}_i$ based on the sample $i$: 
\begin{eqnarray}
\small
\label{eq:ennloss}
\mathcal{L}(f({\bf x}_i|\Theta), {\bf y}_i) 
=\int  \frac{\|{\bf y}_i - {\bf p}_i\|_2^2}{B({\bm \alpha}_i)} \prod_{j =1}^K p_{ij} ^{(\alpha_{ij}-1)} d{\bf p}_i \nonumber \\
=\sum_{j=1}^K (y_{ij}^2 - 2 y_{ij}\mathbb{E}[p_{ij}] +\mathbb{E}[p_{ij}^2]). 
\end{eqnarray}

Eq.~\eqref{eq:ennloss} is based on class labels of training samples. However, it does not directly measure the quality of the predicted Dirichlet distributions such that the uncertainty estimates may not be accurate.

\section{Training ENNs with Wasserstein GAN} \label{sec:training-enn-gen-model}
 Given the various characteristics of uncertainty based on SL, we propose a novel model that combines ENNs and WGAN to quantify multiple types of uncertainty (i.e., vacuity and dissonance) and solving classification tasks.

\subsection{Regularized ENNs} \label{subsec:regularized-enn}
Given a set of samples  $\mathcal{D} = \{({\bf x}_1, {\bf y}_1), \cdots, ({\bf x}_N, {\bf y}_N)\}$, 
let $P_{out}({\bf x}, {\bf y})$ and $P_{in}({\bf x}, {\bf y})$ be the distributions of the OOD and ID samples respectively.
We propose a training method using a regularized ENN to minimize the following loss function over the parameters $\Theta$ of the model's function $f$:
\begin{eqnarray}
\small
\label{eq:regularize_enn}
\mathcal{L}(\Theta) = \mathbb{E}_{{\bf x, \bf y}\sim P_{in}({\bf x, \bf y})}[\mathcal{L}(f({\bf x}|\Theta), {\bf y}) ] \\\nonumber
- \beta \mathbb{E}_{{\bf \hat{x}} \sim P_{out}({\bf  \hat{x}})}[\text{Vac}(f({\bf \hat{x}}|\Theta))].
\end{eqnarray}

The first item (Eq.~\eqref{eq:ennloss}) ensures a good estimation of the class probabilities of the ID samples. Since it assigns large confidence on training samples during the classification process, it also contributes to reducing the vacuity of ID samples. The second item is to increase the vacuity estimation from the model on OOD samples. $\beta$ is the trade-off parameter. Therefore, minimizing Eq.~\eqref{eq:regularize_enn} is to achieve high classification accuracy, low vacuity output for ID samples and high vacuity output for OOD samples. 
To ensure  the  model's  generalization to the whole data space, 
the choice of effective $P_{out}$ is important.
While some methods~\cite{lee2017training,hein2019relu,sensoyuncertainty} only use close or adversarial samples, we found that both close and far-away samples are equally important. 
Instead of using hand-picked auxiliary dataset $P_{out}$ which requires a lot of tuning~\cite{zhao2019quantifying}, we used generative models to provide sufficient various OOD samples.

\subsection{Wasserstein Generators for OOD}

We chose WGAN~\cite{arjovsky2017wasserstein} as our generators because (i) it provides higher stability than original GAN~\cite{goodfellow2014generative}; and (ii) it offers a meaningful loss metric by leveraging the Wasserstein distance, which can measure the distance between the generated samples and the ID region.

WGAN consists of two main components: Discriminator $D$ and generator $G$. $G$ maps input latent variable $z$ into generator output $G(z)$ where $D$ represents the probability of input sample ${\bf x}$ coming from ID. The objective function is to recover $P_{in}({\bf x})$ from $G$. WGAN uses the Wasserstein distance instead of the original divergence in the GAN's loss function, which is considered as continuous and differentiable for optimization. The Wasserstein distance $dist(p,q)$ between two distributions $p$ and $q$ is informally defined as the minimum cost of transporting mass to transform $q$ into $p$. Under the Lipschitz constraint, the loss function of WGAN can be written as:
\begin{eqnarray}
\small
\min_G \max_D \mathbb{E}_{{\bf x}\sim P_{in}({\bf x})} \big[D({\bf x}) \big] 
- \mathbb{E}_{\hat{{\bf x}}\sim P_G(z)} \big[D(\hat{{\bf x}}) \big],
\label{eq:wgan_loss}
\end{eqnarray}
where $P_{in}$ is the ID and $P_G$ is the the generated distribution defined by $x = G(z)$ and $z \sim p(z)$, which is usually sampled from uniform or Gaussian noise distribution. WGAN employs weight clipping or gradient penalty (WGAN-GP)~\cite{gulrajani2017improved} to enforce a Lipschitz constraint to keep the training stable.  
We can estimate the Wasserstein distance $dist$ at the step after $D$ updates and before $G$ updates during the alternating training process:
\begin{eqnarray}\small
 \mathbb{E}_{{\bf x}\sim P_{in}({\bf x})} \big[D({\bf x}) \big] 
- \mathbb{E}_{\hat{{\bf x}}\sim P_G(z)} \big[D(\hat{{\bf x}}) \big]. 
\end{eqnarray}

The estimated curve of $dist$ during WGAN training shows high correlation with high visual quality of the generated samples~\cite{arjovsky2017wasserstein}. When training WGAN from the scratch, the initial large distance indicates that the generated samples has very low-quality, showing far-away samples to the ID region.  Through the progress of the training, the distance decreases continuously, which leads to higher sample quality. This implies that the samples are getting close to the ID region.  
Therefore we adopted $dist$ to measure the variety of generated samples, which are used as prior knowledge of our model.

However, original WGAN is designed to generate ID samples. To reinforce $G$  recover OOD $P_{out}$, we propose the following new WGAN loss with uncertainty regularization.
\begin{eqnarray}\small
\label{eq:wgan_reg}
\min_G \max_D \mathbb{E}_{{\bf x}\sim P_{in}({\bf x})} \big[D({\bf x}) \big] - \mathbb{E}_{\hat{{\bf x}}\sim P_G(z)} \big[D(\hat{{\bf x}}) \big] \\
-  \beta \mathbb{E}_{\hat{{\bf x}}\sim P_G(z)} \big[\text{Vac}(f({\bf \hat{\bf x}}|\Theta)) \big]\nonumber , 
\end{eqnarray}
where $\beta$ is a trade-off parameter and $\text{Vac}(f(\hat{\bf x}|\Theta))$ is the uncertainty estimation from a classifier trained on ID.  This regularization item enforces the generated samples to have high vacuity uncertainty. 


\subsection{Jointly Training ENNs and WGAN}


\begin{algorithm}[h]
\small
\caption{Alternating minimization for WGAN and ENN}
\label{alg:algorithm}
\begin{algorithmic}
\Require  {Pretrained ENN with weights $\Theta$, initial $D$'s weights $\omega$, initial $G$'s weights $\theta$. $n_{d}$, $n_{e}$ : The number of iterations of $D$ and ENN per $G$ iteration. $\beta$: the trade-off weight. $m$: the batch size. }
\end{algorithmic}

\begin{algorithmic}[1]

\Repeat
\For{ $i = 1,...,  n_{d} $} 
\State {Sample $  \{{\bf z}^{(i)}\}_{i=1}^m \!\sim\! P_z$ and $  \{{\bf \hat x}^{(i)}\}_{i=1}^m \!\sim\! P_G(z)$}

\State {Update $D$ by descending its gradient (with penalty)}  

\Statex  {\qquad \quad  $\nabla_{\omega} \frac{1}{m}\sum\limits_{i=1}^{m}\big[ D_{\omega}({\bf \hat{x}}^{(i)}) 
         \! - \! D_{\omega}({\bf x}^{(i)}) + grad\_penalty \big]\nonumber $}
\EndFor

\State {Get the approximated Wasserstein distance} 
\Statex  {\qquad \quad  $dist = \frac{1}{m}\sum\limits_{i=1}^{m}\big[ D({\bf x}^{(i)}) \! -\! D({\bf \hat{x}}^{(i)}) \big]\nonumber$}

\State {Update $G$ once by ascending its gradient} 

\Statex {\qquad \quad  $ \nabla_{\theta}\frac{1}{m}\sum\limits_{i=1}^{m}\big[\! D(G_{\theta}({\bf z}^{(i)}))
                + \beta\text{Vac}(f(G_{\theta}({\bf z}^{(i)})|\Theta))
                \big{]}\nonumber$}

\For{ $i = 1,...,  n_{e} $}
\State {Sample $ \{{\bf x}^{(i)},{y}^{(i)} \}_{i=1}^m \!\sim\! P_{in}\nonumber$\nonumber, $\{{\bf z}^{(i)}\}_{i=1}^m \!\sim\! P_z\nonumber$ , $\{{\bf \hat x}^{(i)}\}_{i=1}^m \!\sim\! P_G(z)$}

\State {Update ENN by descending the gradient}

\Statex {\qquad \quad  $\nabla_{\Theta}\frac{1}{m}\sum\limits_{i=1}^{m}\big{[}\mathcal{L}(f({\bf x}^{(i)}|\Theta), y^{(i)} ) \big{]}\nonumber$}
                
\State {Update ENN by ascending the gradient}

\Statex {\qquad \quad  $\nabla_{\Theta}\frac{\beta}{m}\sum\limits_{i=1}^{N}\big[\text{Vac}(f({\bf \hat{x}}^{(i)}|\Theta)) \big]\nonumber$}
\EndFor
\Until{$dist$ convergence}

\end{algorithmic}
\end{algorithm}

To improve the OOD detection accuracy, we jointly trained regularized ENN loss (Eq.~\eqref{eq:regularize_enn}) and WGAN-GP with uncertainty regularization (Eq.~\eqref{eq:wgan_reg}) in Algorithm~\ref{alg:algorithm}. This allows ENNs to utilize various types of OOD samples generated from WGAN. We usually pretrain the ENN classifier for a good classification accuracy to accelerate the training.

Each batch of the OOD samples correspond to different decreasing $dist$. This enables the ENN to utilize a wide range of OOD samples. ENNs improves uncertainty estimation based on OOD samples from $P_{G}$, and $G$ achieves a better OOD quality due to uncertainty estimation from ENNs simultaneously. We stop the training when $dist$ converges in case the ENN may forget the effect of previous far-away OOD samples from ID regions.  Fig~\ref{fig: was_distance} illustrates the change of  $dist$ and output vacuity during the training process.

\section{Experiments}

We first illustrated the advantage of evidential uncertainty in (1) a synthetic experiment. Then we compared our approach with the recent uncertainty estimation models on (2) predictive uncertainty estimation and (3) adversarial uncertainty estimation. (4) We also investigated the effect of different types of uncertainties on the OOD detection.

\begin{figure}[htbp]
\centering
\includegraphics[width=0.47\textwidth]{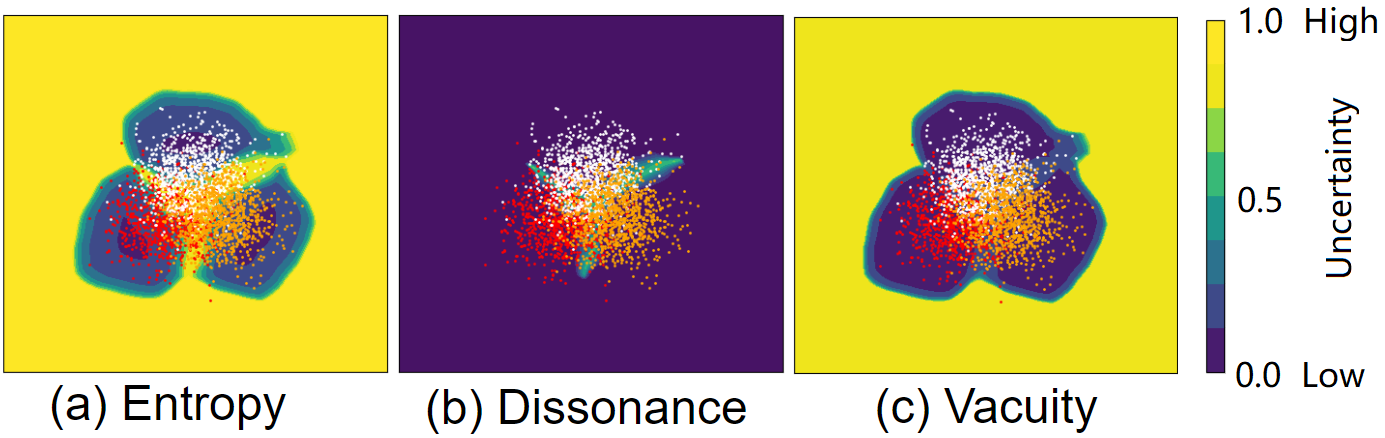}
\caption{A synthetic experiment that shows the benefit of using evidential uncertainties (vacuity, dissonance) over entropy ranged in [0, 1].  Entropy  cannot distinguish ID and OOD samples at class boundaries. 
}\label{fig: syn_exp}
\end{figure}

\subsection{Synthetic}

Fig~\ref{fig: syn_exp} shows three Gaussian distributed classes with equidistant means and tied isotropic variance $\sigma^2=4$ (a large degree of class overlap).
We used our proposed WENN method, a small NN with 2 hidden layer of 500 neurons each was trained on this data. Fig~\ref{fig: syn_exp} demonstrates that entropy and two evidential uncertainties, which are vacuity, dissonance, exhibit distinctive behaviors. 
Entropy is high both in overlapping and far-away regions from training data, which makes it hard to distinguish ID and OOD samples at class boundaries. In contrast, vacuity is low over the whole region of training data while vacuity is high for the outside of the region of training data. This allows the ID region to be clearly distinguished from the OOD region.  In addition, high dissonance is observed over decision boundary which indicates high chances of misclassification.

\begin{figure*}[htb]
    \centering
    \begin{subfigure}[t]{0.33\linewidth}
        \centering
        \includegraphics[width=\linewidth]{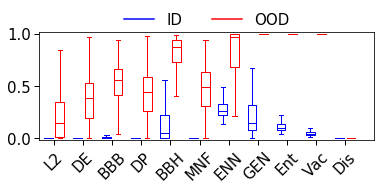}
        \caption{MNIST: ID vs OOD}\label{fig:boxplot_ood_1}
    \end{subfigure}
    \begin{subfigure}[t]{0.33\linewidth}
        \centering
        \includegraphics[width=\linewidth]{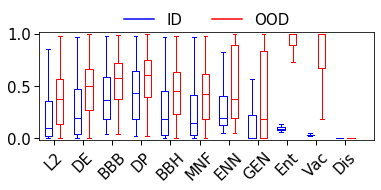}
         \caption{CIFAR10: ID vs OOD}\label{fig:boxplot_ood_2}
    \end{subfigure}
    \begin{subfigure}[t]{0.33\linewidth}
        \centering
        \includegraphics[width=\linewidth]{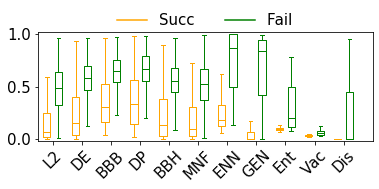}
         \caption{CIFAR10: Successful vs Fail predictions }\label{fig:boxplot_succ_fail}
    \end{subfigure}
    
    \small{
    \caption{Boxplots of predictive uncertainty of different models on ID/OOD testing datasets: (a) MNIST test set (ID) vs notMNIST (OOD); (b) Samples from the first five (ID) vs the last five (OOD) categories of CIFAR10 test set; (c) Successful and failed predictions in CIFAR10 test set (ID). 
    Our model uses entropy (Ent), vacuity (Vac), dissonance (Dis) as a  measure  of  uncertainty, while other models use entropy. }\label{fig:boxplot_ood}
    }
\end{figure*}

\subsection{Predictive Uncertainty Estimation}

\noindent {\bf Comparing Schemes}: We compared our model with the following schemes: 
(i) \textbf{L2} refers to the standard deterministic NNs with softmax output and weight decay; 
(ii) \textbf{DP} uses Dropout, the uncertainty estimation model (i.e., BNNs)~\cite{gal2016dropout};
(iii) \textbf{DE} refers to Deep Ensembles~\cite{lakshminarayanan2017simple}; 
(iv) \textbf{BBB} refers to Bayes by Backprop ~\cite{blundell2015weight};
(v) \textbf{BBH} refers to Bayes by Hypernet~\cite{pawlowski2017implicit}, a Bayesian model based on implicit weight uncertainty;
(vi) \textbf{MNF} refers to the variational approximation based model in~\cite{louizos2017multiplicative};
(vii) \textbf{ENN} uses evidential DL model~\cite{sensoy2018evidential}; 
(viii) \textbf{GEN} combines ENNs and Adversarial Autoencoder~\cite{sensoyuncertainty}; and 
(ix) \textbf{Ent}, \textbf{Vac} and \textbf{Dis}  are the entropy, vacuity and dissonance of our proposed model \textbf{WENN}.
\\

\noindent {\bf Setup}: We followed the same experiments in~\cite{sensoyuncertainty} on MNIST~\cite{lecun1998gradient} and CIFAR10~\cite{krizhevsky2009learning}: 
(1) For the {\bf MNIST dataset}, we used the same LeNet-5 architecture from~\cite{sensoyuncertainty}. We trained the model on MNIST training set and tested on MNIST testing set as ID samples and notMNIST~\cite{notMNIST} as OOD samples; 
and (2) For the {\bf CIFAR10 dataset}, we used ResNet-20~\cite{he2016deep} as a classifier in all the models considered in this work. We trained on the samples for the first five categories \{airplane, automobile, bird, cat, deer\} in the CIFAR10 training set (i.e., ID), while using the other five categories \{ship, truck, dog, frog, horse\} as testing OODs. 
We used the source code of BBH and GEN,  which also contained implementations of other approaches. 
But we changed all the classifiers to the same LeNet-5 and ResNet-20 respectively. 
All the baselines were fairly trained with their default best parameters and we reported the average results. 
For WENN, we set $\beta = 0.1$,  $n_{d} = 2$, $n_{e} =1$, $m =256$, $\mathrm{learning\_rate}=1e^{-4}$ in Algorithm~\ref{alg:algorithm} in all the experiments, which were fine-tuned considering the performance of both the OOD detection and ID classification accuracy. For more details refer to Appendix and our source code
\footnote{\url{https://github.com/snowood1/wenn}}. 
\\

\noindent {\bf Metrics}: Our proposed model used vacuity and dissonance estimated based on Eq.~\eqref{eq:vacuity} and Eq.~\eqref{eq:belief-dissonance-multi}. To be consistent with other works that used entropy as a measure of uncertainty, 
we also compared the predictive entropy over the range of possible entropy $[0, 1]$. 
We used the boxplots to show the distribution of predictive uncertainty.
\\



\noindent {\bf Results}: To evaluate OOD uncertainty qualification, Fig~\ref{fig:boxplot_ood} (a) and (b) show the boxplots of the predictive uncertainty under all models trained with MNIST and CIFAR10 and tested on their corresponding ID and OOD datasets. The ideal model is expected to have a low ID box and a high OOD box, i.e., the model is certain about the ID inputs while totally uncertain about the OOD inputs. To measure ID uncertainty qualification, Fig~\ref{fig:boxplot_ood} (c) shows the boxplots of different models' predictive uncertainty for correct and mis-classified examples in CIFAR10 ID testing set.  
The figure indicates that a standard network is overconfident of any inputs. BBH performs the best among all the Bayesian models on MNIST but fails to give a disparity between ID and OOD on CIFAR10.  ENN and GEN perform well on MNIST. 
However, Fig~\ref{fig:boxplot_ood} (b) and (c) show that they force high uncertainty for mis-classified ID samples the same as OOD samples on CIFAR10. \cite{sensoyuncertainty} admits that ENN and GEN may classify the boundary ID samples as OOD because of their high entropy.
The above results all indicate the limitation of entropy in uncertainty estimation. 

WENN using entropy beats other counterparts in estimating OOD uncertainty because it benefits from our algorithm using vacuity.
However, WENN is more powerful when using vacuity and dissonance to measure OOD and ID uncertainty respectively.
For ID uncertainty, Fig~\ref{fig:boxplot_ood} (c) illustrates that high dissonance implies conflicting evidence, which can result in mis-classification.   
For OOD uncertainty, Fig~\ref{fig:boxplot_ood} (b) and (c) show that all the ID samples, i.e., even the mis-classified  samples, have extreme low vacuity, compared to the high vacuity of OOD samples. However, WENN still assigns medium entropy to boundary ID samples. This is consistent with the synthetic experiment's result, showing the advantage of adopting vacuity in distinguishing boundary ID and OOD samples.

\subsection{Adversarial Uncertainty Estimation}
\begin{figure*}[htb]
    \small{
    \centering
    \begin{minipage}[t]{0.48\linewidth}
        \centering
        \includegraphics[width=0.88\textwidth]{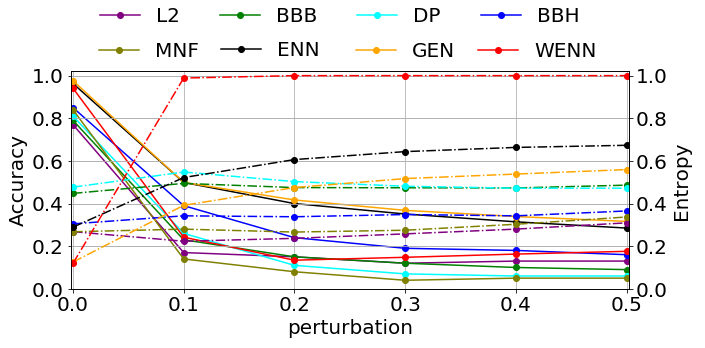}
        \caption{Accuracy (solid) vs entropy (dashed) as a function of
        the adversarial perturbation $\epsilon$ on CIFAR10.} 
        \label{fig:adv}
    \end{minipage}
    \hfill
    \begin{minipage}[t]{0.48\linewidth}
        \centering
        \includegraphics[width=0.88\textwidth]{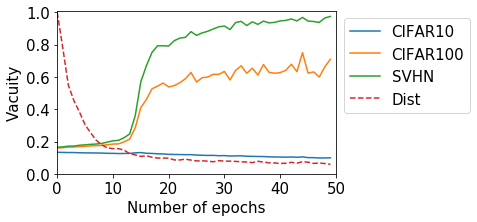}
        \caption{The change of normalized Wasserstein distance and the vacuity of ID/OOD samples when trained on CIFAR10.}
        \label{fig: was_distance}
    \end{minipage}
    }
\end{figure*}

We also evaluated these models on CIFAR10 using adversarial examples generated by FGSM \cite{goodfellow2014explaining} with different perturbation values $\epsilon \!\in \![0, 0.5]$.  DE is excluded because it is trained on adversarial samples.  Fig~\ref{fig:adv} shows that as $\epsilon$ increases, WENN's accuracy immediately drops to random and the uncertainty simultaneously increases to maximum entropy, i.e., WENN will assign the highest uncertainty with the inputs if it can't make easy predictions. It knows what it doesn't know and never becomes overconfident. We observe the same behaviors for MNIST dataset (in Appendix).

\begin{table*}[htbp]
\centering\fontsize{9}{9}\selectfont
\begin{tabular}{l|l|cccccccccc} 
\toprule
\textbf{ID}\tnote{*} & \textbf{ OOD}& \textbf{MSP} & \textbf{ENN} & \textbf{GEN} &\textbf{CCC} & \textbf{ODIN}& \textbf{ACET} &\textbf{OE} &\textbf{CCU} 
& \textbf{Ent} & \textbf{Vac}\\

\midrule
\multirow{2}{*}{\rotatebox{0}{FMNIST}} 
 &MNIST     & 96.9 & 90.1 & 96.2 & 99.9 & 99.2 & 95.6 & 92.0 & 96.2 &\textbf{100.0}  &\textbf{100.0} \\
 &notMNIST  & 87.5 & 87.8 & 93.6 & 96.4 & 90.2 & 92.4 & 93.0 & 96.7 & 99.9  &\textbf{100.0}  \\
 &Uniform  & 93.0 & 91.6 & 96.9 & 95.4 & 94.9 & \textbf{100.0} & 99.3 &\textbf{100.0} & 99.9 &\textbf{100.0} \\
\midrule
\multirow{3}{*}{\rotatebox{0}{CIFAR10}} 
 &CIFAR100  & 86.3 & 75.0 & 84.0 & 84.0 & 87.1 & 85.2 & 86.0 & 92.5  & 98.6 & \textbf{99.5}  \\
 &SVHN      & 88.9 & 78.6 & 85.4 & 80.5 & 85.1 & 89.6 & 92.1 & 98.9  & \textbf{100.0} & \textbf{100.0}   \\
 &LSUN\_CR  & 88.8 & 64.4 & 98.0 & 99.7 & 92.8 & 89.1 & 92.7 & 98.6  &\textbf{100.0}  &\textbf{100.0}  \\
 &Uniform   & 93.8 & 84.6 & 82.4 & 82.4 & 99.3 & \textbf{100.0} & \textbf{100.0} & \textbf{100.0} & \textbf{100.0}  &\textbf{100.0} \\ 
\midrule
\multirow{3}{*}{\rotatebox{0}{SVHN}} 
 &CIFAR10   & 95.2  & 52.9 & 50.2 & 98.6            &95.8   & 96.3           &\textbf{100.0}  &\textbf{100.0}  & 99.3  &\textbf{100.0}  \\
 &CIFAR100  & 94.9  & 51.8 & 51.0 & 98.2            &95.3   & 95.6           &\textbf{100.0}  &\textbf{100.0}  & 99.3  & \textbf{100.0}  \\
 &LSUN\_CR  & 94.9  & 55.5 & 53.9 & \textbf{100.0}  & 95.6  & 97.0           &\textbf{100.0}  & \textbf{100.0} & 99.8  & 99.9  \\
 &Uniform    & 95.8	& 53.9 & 53.2 & \textbf{100.0}  & 96.6  & \textbf{100.0} & \textbf{100.0} &	\textbf{100.0} & 99.8  & \textbf{100.0}\\ 
\bottomrule
\end{tabular}

\caption{\normalsize{AUROC for OOD detection. ID: in-distribution samples, OOD: out-of-distribution samples.}}

\label{table:au}

\end{table*}

\subsection{Out-of-Distribution Detection}

\noindent {\bf Comparing Schemes}: 
We compared with several recent methods specifically designed for OOD detection, together with uncertainty models ENN and GEN: 
(i) \textbf{MSP} refers to maximum softmax probability, 
a baseline of OOD detection in~\cite{hendrycks2016baseline};
(ii) \textbf{CCC}~\cite{lee2017training} 
uses GAN to generate boundary OOD samples as regularizers;
(iii) \textbf{ODIN} calibrates the estimated confidence by scaling the logits before softmax layers~\cite{liang2017enhancing};
(iv) \textbf{ACET} uses adversarial examples to enhance the confidence \cite{hein2019relu};
(v) \textbf{OE} refers to Outlier Exposure~\cite{hendrycks2018deep} that enforces uniform confidence on 80 million Tiny ImageNet \cite{torralba200880};
(vi) \textbf{CCU} integrates Gaussian mixture models in OOD detection DL models~\cite{meinke2019towards}; and
(vii) \textbf{Ent} and \textbf{Vac} refer to our \textbf{WENN} model using entropy or vacuity as scores.
\\

\noindent {\bf Setup}: We used FashionMNIST~\cite{xiao2017}, notMNIST, CIFAR10, CIFAR100~\cite{krizhevsky2009learning},  SVHN~\cite{netzer2011reading}, the classroom class of LSUN (i.e., LSUN\_CR)~\cite{yu2015lsun} and uniform noise as ID or OOD datasets.  We used the source code in \cite{meinke2019towards} which contained implementations of other baselines, but we used ResNet-20 for all the models except CCC. We used VGG-13~\cite{simonyan2014very} for CCC because we couldn't achieve an acceptable accuracy using ResNet-20. And ODIN, OE and CCU were directly trained or calibrated on the Tiny ImageNet. Other settings were the same as the previous uncertainty estimation experiments. 
\\

\noindent {\bf Metrics}: Our model uses vacuity to distinguish between ID and OOD samples. ENN and GEN use the entropy of the predictive probabilities as recommended in their papers. The other methods use their own OOD scores. We use area under the ROC (AUROC) curves to evaluate the performance of different type of uncertainty.
\\

\noindent {\bf Results}: 
Table~\ref{table:au} shows the AUROC curves performance of different approaches.  WENN's vacuity beats all the other uncertainty scores, including its own entropy.
ENN and GEN are not originally designed for OOD detection because they assign the same high entropy to mis-classified ID samples as OOD.
CCC doesn't generalize well and it lacks scalability to recent deep architectures like ResNet to ensure a better classification accuracy. 
The result of ACET proves that the effect of using purely close adversarial examples is limited.  \cite{hein2019relu} admits that ACET will yield high-confidence predictions far away from the training data. 
ODIN, OE and CCU are directly trained or tuned using a large auxiliary dataset which should contain both far-away and close samples.
The outperformance of WENN indicates that our algorithm using vacuity can generate and utilize sufficient OOD samples more effectively.

To further explain how our model generates and utilizes variable OOD samples, Fig~\ref{fig: was_distance} illustrates the alternating optimization process when the model is trained on CIFAR10 training set. The initial ENN classifier is overconfident and assigns arbitrary inputs with low vacuity 0.2. As the Wasserstein distance decreases gradually, implying that the generated samples keep moving from far-away to closer to the ID region, the model learns to output low vacuity on ID samples from CIFAR10 testing set and high vacuity on OOD samples from CIFAR100 and SVHN. 
The output vacuity of CIFAR100 is lower than that of SVHN. 
This indicates vacuity is a reasonable uncertainty metric 
because CIFAR100 is often considered as near-distribution outliers of CIFAR10.
However, the medium vacuity of CIFAR100 is still good enough for perfect classification.

\section{Conclusion} \label{sec:conclusion-future-work}

We proposed a novel DL model, called WENN that combines ENNs with WGAN, to jointly train a model with prior knowledge of a certain class (i.e., high vacuity OOD samples). Via extensive experiments based on both synthetic and real datasets, we proved that: (1) vacuity can distinguish boundary samples from OOD samples; 
(2) the proposed model with vacuity regularization can produce and utilize various types of OOD samples successfully. 
Our model achieved the state of the art performance in both uncertainty estimation and OOD detection benchmarks.





\section{Acknowledgments}
This work is supported by NSF awards IIS-1815696 and IIS-1750911.

\bibliography{reference}

\end{document}